# A survey and taxonomy of methods interpreting random forest models

Maissae Haddouchi and Abdelaziz Berrado

**Maissae Haddouchi**

Mohammed V University in Rabat, Ecole Mohammadia d'Ingénieurs (EMI), Rabat, Morocco. E-mail: maissaehaddouchi@research.emi.ac.ma
ORCID iD: 0000−0002−4851−7979.

**Abdelaziz Berrado**

Mohammed V University in Rabat, Ecole Mohammadia d'Ingénieurs (EMI), Rabat, Morocco. E-mail: berrado@emi.ac.ma

## Conflict of Interest

On behalf of all authors, the corresponding author states that there is no conflict of interest.

## Abstract

The interpretability of random forest (RF) models is a research topic of growing interest in the machine learning (ML) community. In the state of the art, RF is considered a powerful learning ensemble given its predictive performance, flexibility, and ease of use. Furthermore, the inner process of the RF model is understandable because it uses an intuitive and intelligible approach for building the RF decision tree ensemble. However, the RF resulting model is regarded as a "black box" because of its numerous deep decision trees. Gaining visibility over the entire process that induces the final decisions by exploring each decision tree is complicated, if not impossible. This complexity limits the acceptance and implementation of RF models in several fields of application.

Several papers have tackled the interpretation of RF models. This paper aims to provide an extensive review of methods used in the literature to interpret RF resulting models. We have analyzed these methods and classified them based on different axes. Although this review is not exhaustive, it provides a taxonomy of various techniques that should guide users in choosing the most appropriate tools for interpreting RF models, depending on the interpretability aspects sought. It should also be valuable for researchers who aim to focus their work on the interpretability of RF or ML black boxes in general.

## 1. INTRODUCTION

The growing progress in computational capabilities and recent advances in computer science developments have made machine learning (ML) algorithms a new kind of powerful "brains" capable of processing massive data and learning the predictive structure of problems in a way that humans cannot. The operational use of these algorithms solves various daily life problems and influences the decision-making process in several critical areas. However, the use of ML algorithms is expected not only to produce accurate predictions but also to provide insights into the predictive structure of the data (Louppe, 2014).

For users (decision makers, analysts, experts, project stakeholders, etc.), the interpretability of ML models is often as important as their predictive performance. When building an ML project, the team needs to understand and trust the ML model, validate the knowledge it provides, and feel reassured during its deployment. The model should be proven faithful and accurate through intelligible explanations. The project team also needs to identify the model's strengths and weaknesses to monitor and consider them when applying ML decisions or leading analysis when problems occur. Moreover, the aim of using ML models in scientific disciplines is often to discover new knowledge that can help experts in the field better understand the phenomenon under study, and interact with the parameters that influence it.

Furthermore, owing to the ethical, moral, or legal impact that an ML decision can produce on one's life, the ML model should argue its decisions through comprehensible explanations. In addition, ML models are constrained in many regulated fields of application to be interpretable so as to provide explanations regarding individual contestations (Goodman and Flaxman, 2017).

The topic of interpretability and explainability of ML models is increasingly gaining importance in the academic community and becoming a crucial aspect for the public acceptance of ML models. Interpretability conventionally refers to the degree to which a human can understand the model decision structure. The terms "interpretability" and "explainability" are generally used synonymously in the literature, although the ML community seems to prefer the former (Preece, 2018; Emmert-Streib et al., 2020; Adadi and Berrada, 2018; Hakkoum et al., 2022). In the rest of this paper, we will use the two terms interchangeably.

In practical application, an algorithm that does not provide, for nonexperts in ML, enough information about the learner process and the learned model can be merely discarded in favor of less accurate and more interpretable approaches (Haddouchi and Berrado, 2018). In the report published in Kaggle's 2021 edition of the State of Machine Learning and Data Science, the most commonly used algorithms were, as in 2020, linear and logistic regression, followed closely by decision trees and random forests. The linear regression algorithm is considered an interpretable algorithm because it provides a clear explanation and a simple graphical representation of the mathematical basis of the prediction structure. However, the linear regression model is generally biased and does not compete in predictive performance with other more complex ML models. It is possible to improve its performance by including polynomial, exponential, or other terms, but this model tweaking will be done at the expense of interpretability. Decision trees are also well-known for their intelligibility in that they mimic, in some sense, the natural reasoning of humans. The popular classification and regression trees (CART) method (Breiman et al., 1884) greedily splits the descriptive variable space into a number of simple and non-overlapping regions, ending up with a diagram describing the logic behind the decision-making process. Each split is performed by comparing the gain in predictive performance based on all the splitting possibilities over all the descriptive variables. This approach obviously provides a simple interpretation of the model (if shallow tree) but generally suffers from overfitting (thus predictive accuracy) compared to other supervised learning methods.

The random forest (RF) model (Breiman, 2001) overcomes the overfitting problem of the individual decision trees by aggregating multiple deep decision trees on different bootstrapped samples and using a random selection of descriptive variables at each splitting step. Thus, an RF ensemble can be seen as a committee of decision-makers (trees) who make decisions through a consensus. Then extracting and interpreting information shared by different points of view can lead to a solid knowledge about the structure of the model. However, this task may seem difficult to achieve given the number and depth of RF trees.

A key claim of this work is that the issue of interpreting the RF model is scientifically interesting. On the one hand, the building process of the RF model is intelligible for human thinking compared to other black-box models. On the other hand, the RF algorithm has proven its efficiency in many practical problems. In one study (Fawagreh et al., 2014; Biau and Scornet, 2016), the authors exposed and referenced several successful applications of RF models in ecology, medicine, agriculture, astronomy, autopsy, bioinformatics and computational biology, chemoinformatics, traffic and transport planning, and 3D object recognition, just to list a few. Also, the algorithm is user-friendly since its tuning parameters are easily understandable (such as the number of trees, the number of candidate variables for splitting, the number of leaf nodes, etc.). Its default hyperparameters are generally sufficient to provide satisfactory results, which results in time-saving. Moreover, the RF algorithm is versatile enough to deal with small sample sizes or be applied to high-dimensional feature spaces (Biau and Scornet, 2016). It can handle big data via parallelization as well (Chen et al., 2017).

The fact that the RF model is categorized as a black-box model (because of its number and depth of trees) restricts its deployment in many fields of application. Hence, providing insight into this model's predictions becomes necessary for its applicability in the real world. Providing a firm basis from the literature about the different works proposing methods to interpret the RF resulting model should guide, in practice, the selection of the most useful tools for the interpretation and deep analysis of the model results, with regard to the interpretability aspects sought. This should also be useful for researchers who focus their studies on the interpretability of RF models or ML in general.

Different methods in the literature look inside the RF model. The increasing number of publications on this topic produces extensive knowledge from various perspectives. The existing reviews on explainable models examine several classes of methods (generally focusing on neural network models). Two works are particularly interested in the review of methods interpreting RF models (Haddouchi and Berrado, 2019; Aria et al., 2021). The two papers present a comprehensive background of the literature on this topic, following a traditional (narrative) review. They essentially analyze the methodological aspects based on the taxonomy proposed by Haddouchi and Berrado (2019). In addition, the later paper (Aria et al., 2021) presents a comparative analysis between two interpretative methods over different benchmarking datasets.

This paper, in contrast, provides an in-depth review of RF interpretation methods following a systematic methodology. It rigorously reviews these methods, providing a bibliographic analysis of the surveyed papers and a literature classification of the different techniques based on different axes (stage of explanation, objective of interpretability, type of the problem solved, its input and output formats, methodology used for providing explanations, and programming language used in the implementation of the method). To achieve the aim of this study, we have set four main objectives:

- To study the evolution, during the last two decades, of the number of methods interpreting RF

models and to check whether there is any tendency or patterns that organize the reviewed papers.

- To conduct a linguistic analysis of the most occurrent terms in the surveyed papers to bring out information that can aid in the classification of the studied methods.
- To set up a classification grid that considers the different aspects that come into play when choosing and exploiting an interpretative method.
- To establish a taxonomy of the methods reviewed and present the distribution of these methods according to the different classification criteria.
- To analyze the usability of these methods in practical applications

The remainder of this paper is organized as follows. In Section 2, we describe the review methodology followed. In Section 3, we present the bibliographic analysis of the reviewed papers. In Section 4, we report the literature taxonomy and discuss the distribution of the surveyed methods according to the different classification criteria. In Section 5, we discuss usability studies related to RF interpretative methods. Finally, in Section 6, we present the conclusion of this work.

## 2. REVIEW METHODOLOGY

We conducted a thorough literature review by examining journal papers from five academic databases publishing works in the computer science domain: ScienceDirect, SCOPUS, ACM Digital Library, IEEExplore, and the dblp computer science bibliography, in addition to well-cited preprints published on the arXiv database. We have used a keyword-based search to select relevant papers. It consisted of searching for the association of the "interpret*" or "explain*" terms with the set of terms "random forest". The search was done within the title, abstract, and keywords search fields (except for the dblp database, where the search was done within the text field). The literature search was refined by limiting the literature search's timeframe to the period from 2001 to 2021 and the language to "English". In addition, the search was narrowed to the computer science domain in each of the SCOPUS and arXiv databases. The extraction of the results was done between 16/02/2022 and 18/02/2022.

At first, a total of 974 papers were selected. After removing redundant papers, a total of 751 papers were screened based on the titles and abstracts to determine relevant articles for further analysis. 73 papers were then considered for text screening and verification of the inclusion and exclusion criteria. We considered only papers that propose original methods interpreting the RF model that can be applied to large purposes. 10 papers applying existing interpretative methods to specific issues and 4 papers reporting methods tailored to specific problems were excluded. 7 other papers were excluded for being oriented toward prediction improvement, feature selection, or missing values processing. Finally, 3 papers were removed because they were based on a modified version of RF, 4 papers were not considered because they do not include an illustration of the interpretable explanations for RF models, and 2 papers were excluded for being reduced versions of other considered papers. A list of 43 selected papers was then completed with 16 additional papers based on exploring the reference and citation lists of the selected papers (snowballing). These 16 papers included 11 journal papers and 5 reports (approved by a scientific committee).

The final list included 59 papers. Figure 1 shows a step-by-step flow diagram of this selection process. We provide in the following section a bibliographic analysis and a literature classification based on these 59 papers.

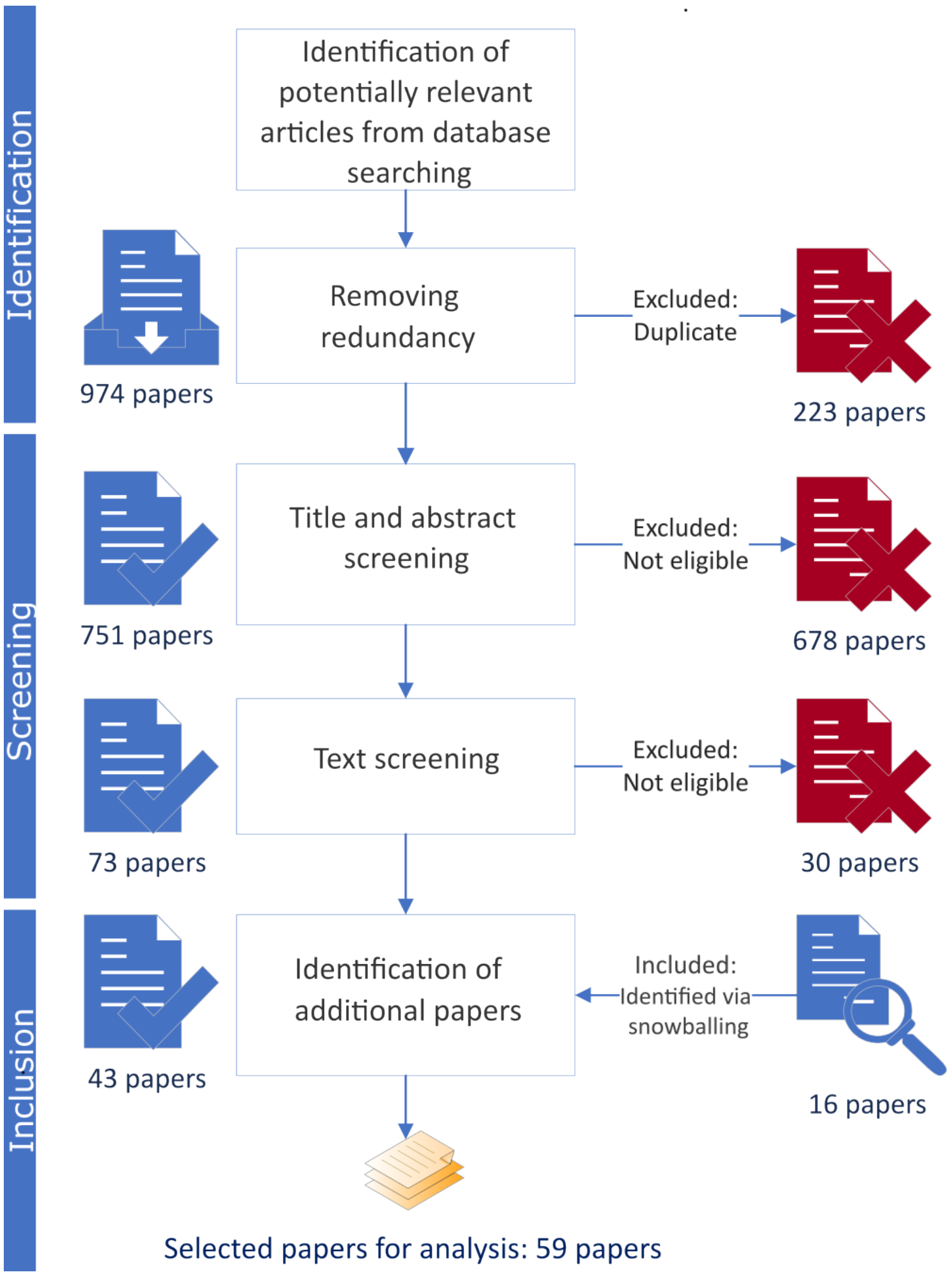

**Figure 1:** Flowchart outlining the selection process of the review of methods interpreting random forest models

## 3. BIBLIOGRAPHIC ANALYSIS

Figure 2 reports the evolution of the number of publications per year. It shows an increasing interest in the research community on the topic of interpreting/explaining RF models, especially in the last 5 years. A similar trend was reported by Adadi and Berrada (2018) regarding XAI in general. The rising interest in the topic of interpretability/explainability in ML is the consequence of the increasing improvement of the ML model's efficiency in solving real-world issues that affect the decision-making process in various critical areas; without being able to provide sufficient information about the key parameters that explain their results (Adadi and Berrada, 2018). However, the more critical an ML model decision is, the more important it is for the model to explain its results. It is therefore foreseeable that the development of more interpretable models will be as important as the development of more accurate models over the next few years. Figure 3 reports the number of citations per paper, clustered based on the paper categories: peer-reviewed journal article, preprint, or report. The citation counts were extracted from Google Scholar on June 19, 2022. The most cited reference is the paper of Liaw and Wiener (2007) (Corpus ID: 3093707 in Semantic Scholar), which introduced two extra pieces of information produced by the RF model: the measure of variable importance and the RF proximity measure. Figure 3 also shows that the highly cited papers are published as preprints and that there was a rising interest in publishing methods interpreting the RF model during the last 3 years.

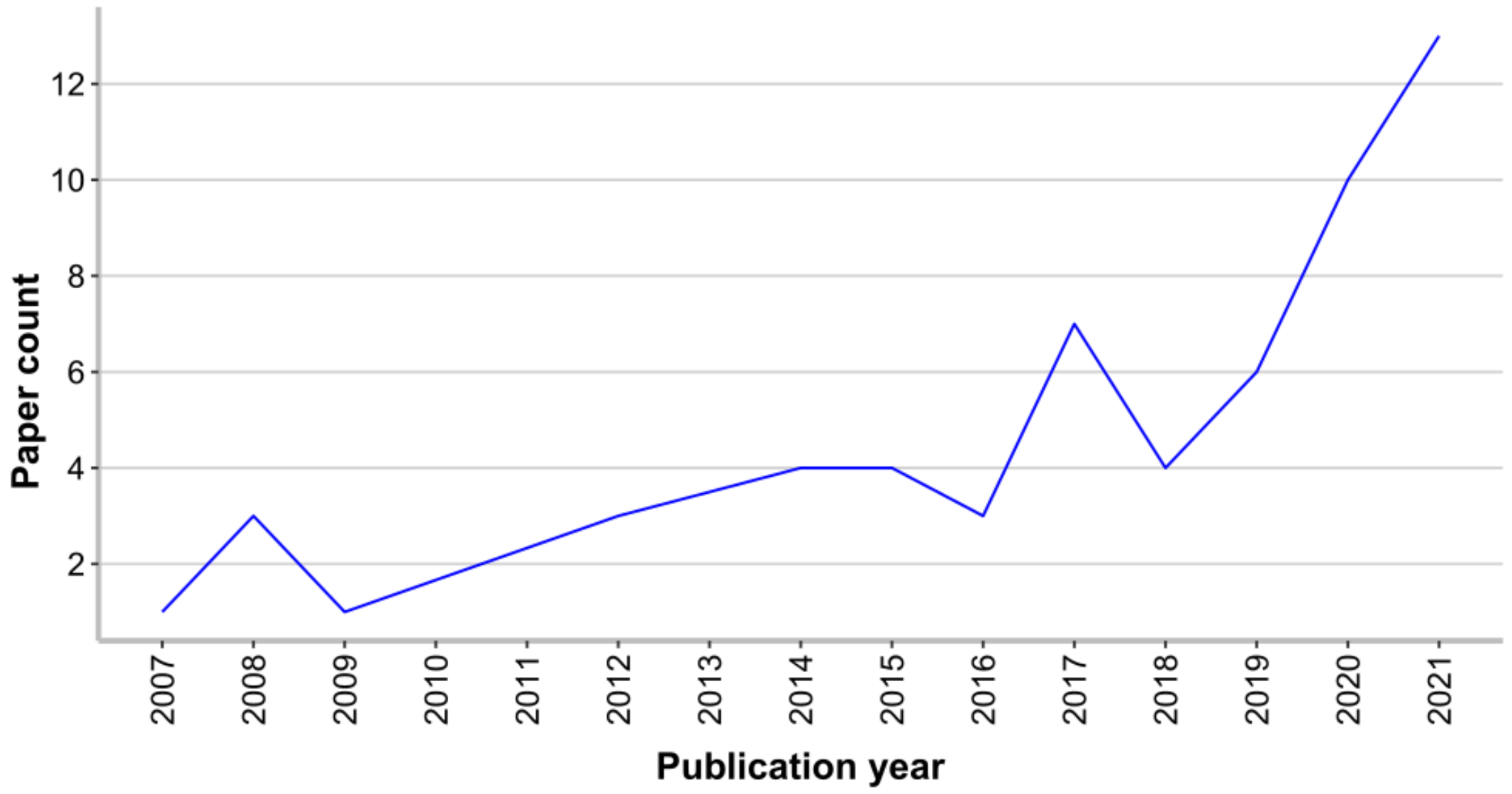

**Figure 2:** Evolution of the number of publications interpreting random forest models per year

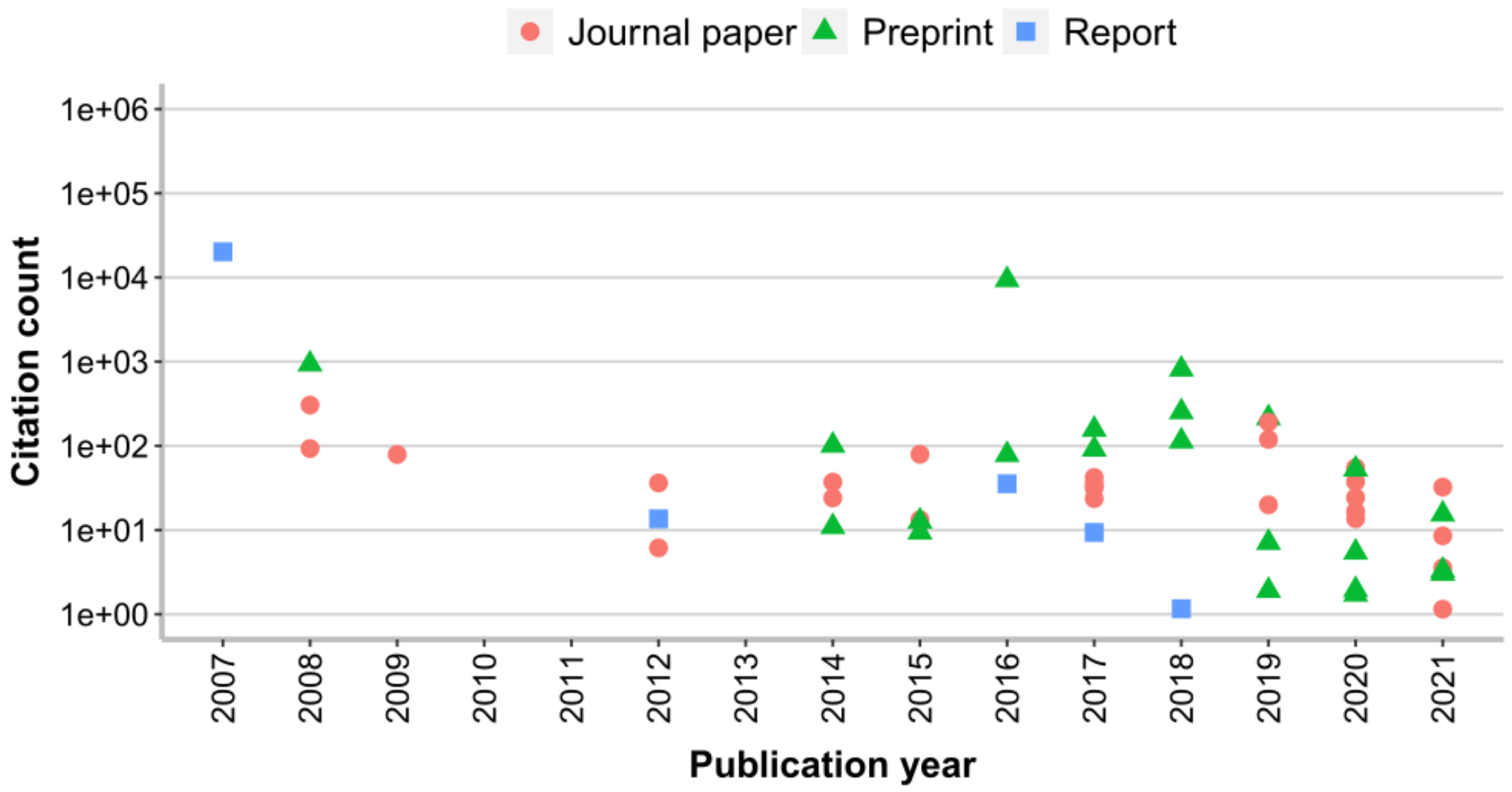

**Figure 3:** Number of citations per paper, clustered based on the paper categories (papers interpreting random forest models)

Figure 4 provides a literature map of authors where the nodes represent the papers and the edges represent their citation relationships (the map was plotted using the Litmaps[1] research platform). The nodes are sized by citation count. This map shows how the papers relate to one another in an ordered timeline (papers published earlier on the left and the latest papers on the right). The most connected papers are, as expected, highly cited papers such as that of Liaw and Wiener (2007). The connection network is globally dense, especially since 2014. Also, we do not notice any apparent clustering in this network. This finding suggests that there is no confirmed direction for this research topic and that most papers considered various previous techniques proposed in the literature.

We performed a linguistic search based on titles and abstracts to discover relevant terms related to RF interpretability. This analysis aims to identify some main concepts that could help classify the existing literature. We provide in Figure 5 a term co-occurrence map based on title and abstract data in the surveyed papers. This map was produced using the VOSviewer[2] software, which scanned the surveyed publications' text data and allowed filtering of irrelevant terms. We have filtered general terms such as "method", "approach", "new", and "result" and technical terms such as the names of the cited algorithms, including "deep learning", "decision tree", "intrees", etc.

[1] https://www.litmaps.com/

[2] https://www.vosviewer.com//

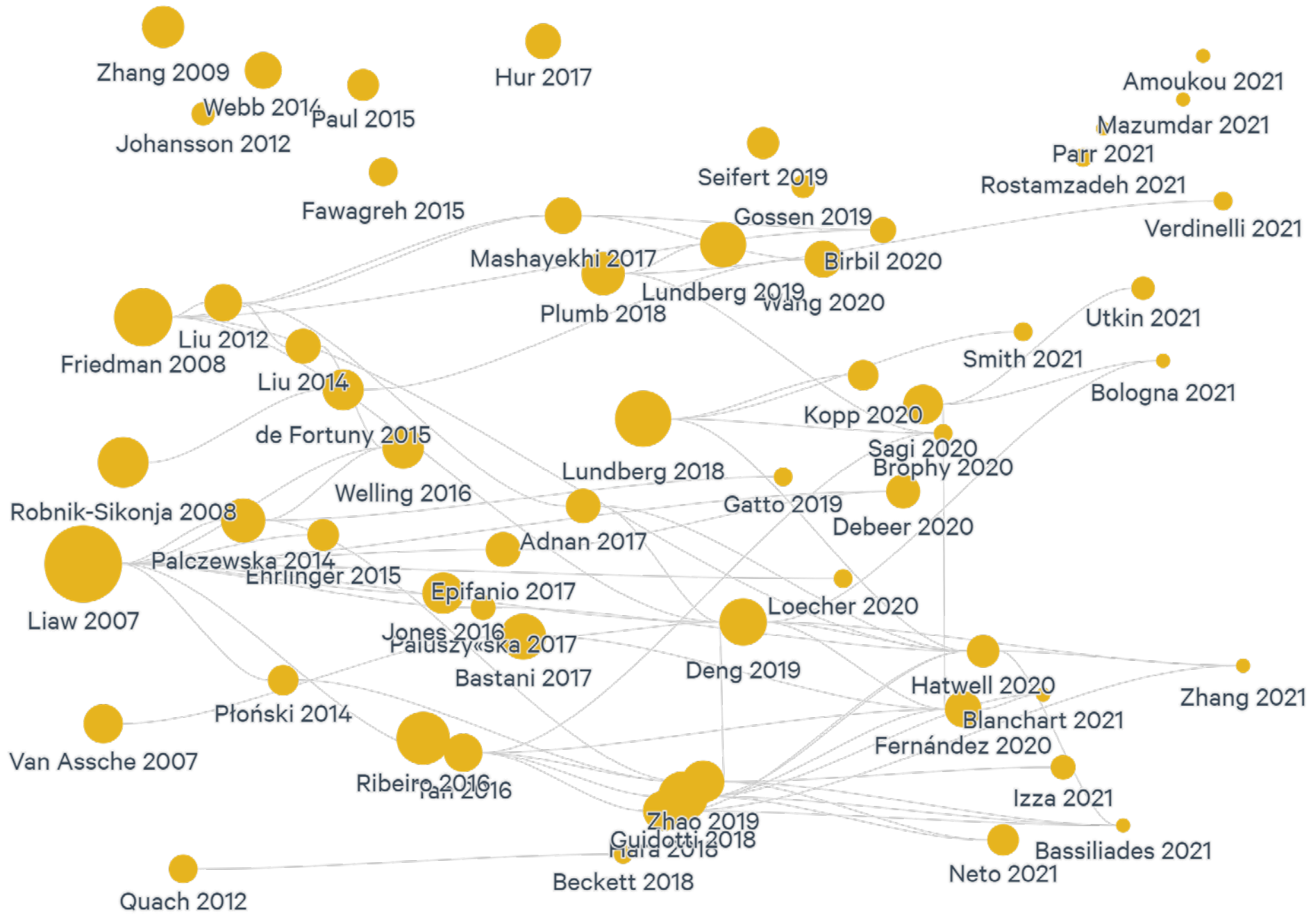

**Figure 4:** Literature map of authors where the nodes represent the papers and the edges represent their citation relationships (for papers interpreting random forest models)

In Figure 5, the nodes represent the relevant terms, and the edges represent their co-occurrence relationship. The nodes are sized by occurrence count. VOSviewer uses modularity-based clustering to identify sets of closely related terms. A detailed description of VOSviewer processing can be found in the paper published by van Eck and Waltman (2010). This map highlights important concepts related to RF interpretability, such as global and local explanation, and surrogate models. It also clusters the methods into different classes of methods, such as rule-based methods, methods producing sub forests, methods based on features' importance, methods based on the similarity among instances, and methods based on interactive visualization. In the following, we propose a literature classification of the surveyed papers based on the concepts above and other additional features. We classified these methods according to seven axes (stage, objective, methodology, issue type, input, output, and code language) to present a comprehensive and holistic analysis of the RF interpretability literature.

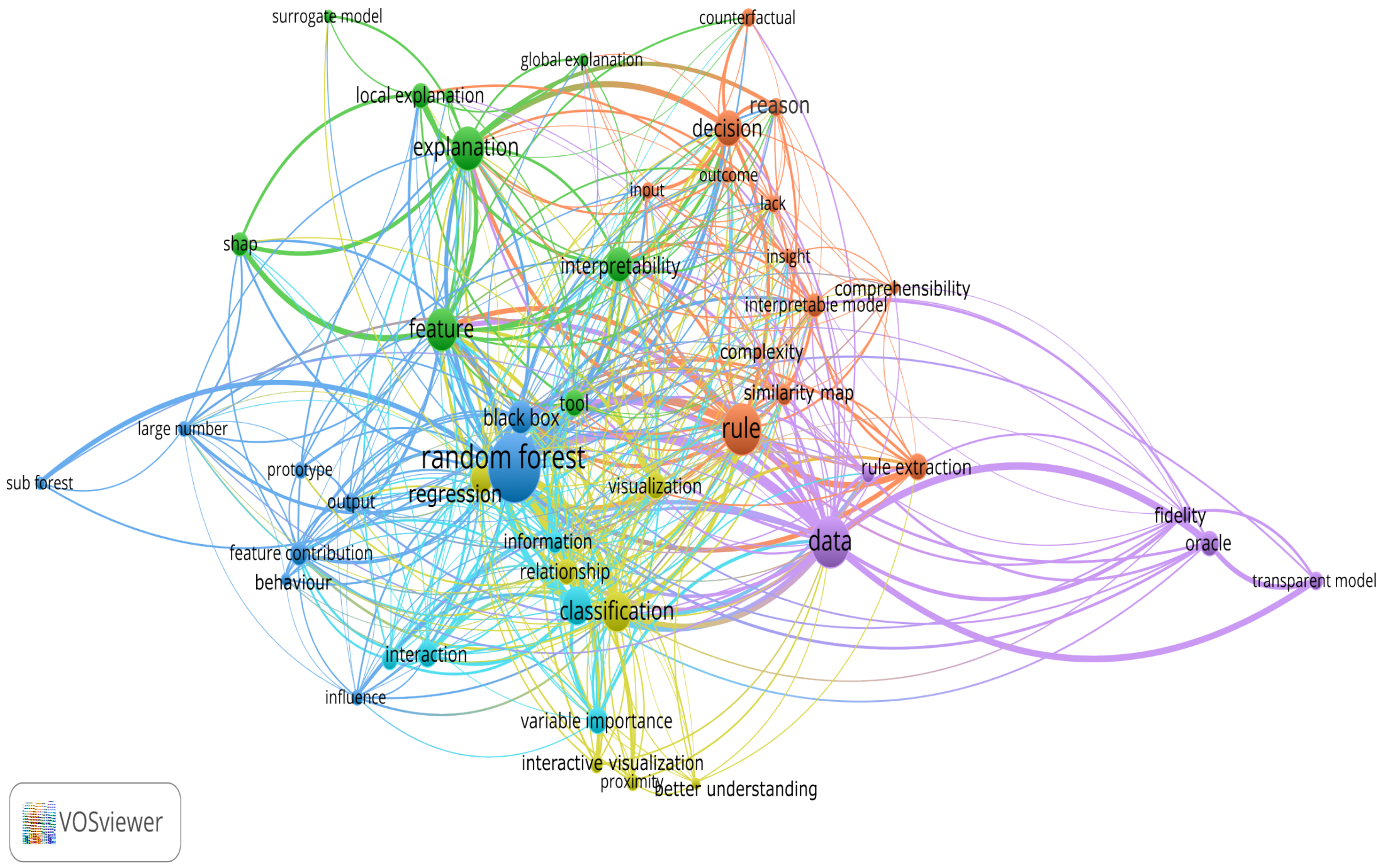

**Figure 5:** Term co-occurrence map based on title and abstract data in the surveyed papers (for papers interpreting random forest models)

## 4. LITERATURE CLASSIFICATION

According to the XAI reviews provided by Vilone and Longo (2020) and Islam et al. (2022), XAI methods can be classified based on the stage of explanation, the type of the problem solved, and its input and output formats. Considering the surveyed articles reported in Section 3 and inspired by the term co-occurrence map (Figure 5), we have added three additional criteria: the objective of interpretability, the methodology used for providing the explanation, and the implementation language used in the implementation of the method. Table 1 lists the methods used in this classification. The name, abbreviation, and reference are provided for each one. The classification details about each method are not provided in this paper because of space restrictions. Instead, readers can consult the GitHub repository reserved for this study, for further information (see Section Declarations).

Figure 6 summarizes the distribution of the surveyed methods over the classification criteria. A detailed analysis of this distribution is provided in the following sections.

**Table 1** List of the selected methods for classification

| Name | Name abrev. | Ref. |
|---|---|---|
| Explaining Classifications for Individual Instances | explainVis | (Robnik-Sikonja and Kononenko, 2008) |
| Combined Rule Extraction and Feature Elimination | CRF | (Liu et al., 2012) |
| Not mentioned | joha12 | (Johansson et al., 2012) |
| Not mentioned | webb14 | (Webb et al., 2014) |
| Feature Contributions for RF Classification | rfFC | (Palczewska et al., 2014) |
| Not mentioned | liu14 | (Liu et al., 2014) |
| Self-Organising Maps | SOM | (Płonski and Zaremba, 2014) |
| Active Learning-Based Pedagogical Rule Extraction | ALPA | (Junque De Fortuny and Martens, 2015) |
| Local Outlier Factor | LOFB-DRF | (Fawagreh et al., 2015) |
| Not mentioned | paul15 | (Paul and Dupont, 2015) |
| Local Interpretable Model-Agnostic Explanations | LIME | (Ribeiro et al., 2016) |
| Shapely Values Contribution | SVC | (Hur et al., 2017) |
| Model Extraction | Modelextraction | (Bastani et al., 2017) |
| Intervention in Prediction Measure | IPM | (Epifanio, 2017) |
| SHapley Additive exPlanation | SHAP | (Lundberg et al., 2018) |
| Model Agnostic Supervised Local Explanations | MAPLE | (Plumb et al., 2019) |
| SHapley Additive exPlanation | ADD-Lib | (Gossen et al., 2019) |
| Tree Explainer | TreeExplainer | (Lundberg et al., 2019) |
| Interpreting Random Forests | iForest | (Zhao et al., 2019) |
| Interpretable Trees | inTrees | (Deng, 2019) |
| Single Sample Feature Importance | SSFI | (Gatto et al., 2019) |
| Surrogate Minimal Depth | SMD | (Seifert et al., 2019) |
| Random Forest-based Rule Extraction | IRFRE | (Wang et al., 2020) |
| Explainer | Explainer | (Kopp et al., 2020) |
| Collection of High Importance Random Path Snippets | CHIRPS | (Hatwell et al., 2020) |
| Improved Conditional Permutation Importance | permimp | (Debeer and Strobl, 2020) |
| Forest Interpretable Tree | forest_based_tree | (Sagi and Rokach, 2020) |
| Explainable AI for Trees | rfVarImpOOB | (Loecher, 2020) |
| Random Forest Optimal Counterfactual Set Extractor | RF-OCSE | (Fernandez et al., 2020) |
| Rule Covering for Interpretation and Boosting | MIRCO | (Birbil et al., 2020) |
| Tree-Ensemble Representer-Point Explanations | TREX | (Brophy and Lowd, 2020) |
| Not mentioned | smit21 | (Smith and Alvarez, 2021) |
| Discretized Interpretable Multi-Layer Perceptron | RF-DIMLP | (Bologna, 2021) |
| Not mentioned | blan21 | (Blanchart, 2021) |
| Conclusive Local Interpretation Rules for Random Forests | LionForests | (Mollas et al., 2021) |
| Consistent Sufficient Explanations and Minimal Local Rules | ACV | (Amoukou and Brunel, 2021) |
| Ensembles of Random SHAPs | ER-SHAP | (Utkin and Konstantinov, 2021) |
| Explainable Matrix | ExMatrix | (M. P. Neto and F. V. Paulovich, 2021) |
| Not mentioned | verd21 | (Verdinelli and Wasserman, 2021) |
| Random Forests Explanations | RFxpl | (Izza and Marques-Silva, 2021) |
| Partial Dependence through Stratification | StratPD | (Parr and Wilson, 2021) |
| Random Forest Similarity Map | RFMap | (Mazumdar et al., 2021) |
| Visual Analytics for Identifying Feature Groups | VERONICA | (Rostamzadeh et al., 2021) |
| Knowledge Discovery from Decision Forests | ForEx | (Adnan and Islam, 2017) |
| Variable Importance/ Partial Dependency /Proximy Matrix | VIP+PDP+PM | (Liaw and Wiener, 2007) |
| Interactive Random Forests Plots | irfplot | (Quach, 2012) |
| Exploring a Random Forest for Regression | ggRandomForests | (Ehrlinger, 2015) |
| Exploratory Data Analysis using Random Forests | edarf | (Jones and Linder, 2016) |
| Interactive Visualization for Random Forests | Rfviz | (Beckett, 2018) |
| Random Forest Explainer | randomForestExplainer | (Paluszy, 2017) |
| Tree Space Prototypes | SG and SM | (Tan et al., 2020) |
| Relational Interpretable Single Model | RISM | (Van Assche and Single , 2008) |
| Sub-Forests | sub-forests | (Wang and Zhang, 2009) |
| Forest Floor | forestFloor | (Welling et al., 2016) |
| Local Rule-Based Explanations | LORE | (Guidotti et al., 2018) |
| Rule Extraction Based on Heuristic Search and Sparse Group Lasso Methods | RF+DHC/SGL/MSGL | (Mashayekhi and Gras, 2017) |
| Not mentioned | FAB | (Hara and Hayashi, 2017) |
| Optimal Explanations for Ensemble | OptExplain | (Zhang et al., 2021) |
| Predictive Learning via Rule Ensembles | RuleFit | (Friedman and Popescu, 2008) |

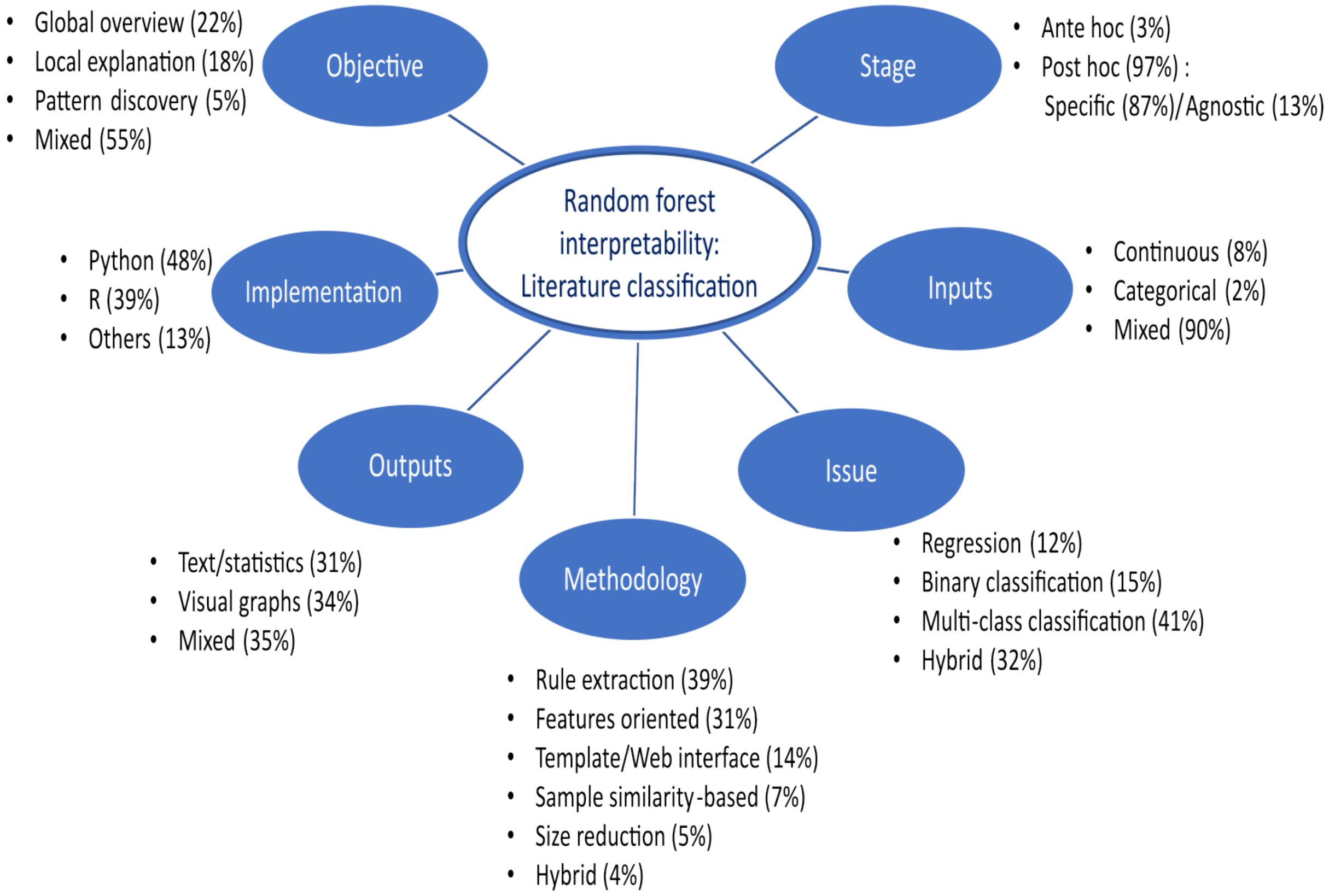

**Figure 6:** Distribution of the surveyed methods over the classification criteria (methods interpreting random forest models)

### 4.1 Taxonomy and distribution analysis of the surveyed papers

#### *4.1.1 Stage*

The stage of explainability designs the time at which the explanation is processed with regard to the processing period of the RF algorithm. The stage can be ante hoc or post hoc (Vilone and Longo, 2020). Around 97% of the surveyed papers propose post hoc methods.

Ante hoc methods consider generating the explanation using the internal processing of the RF model. These explanations concern the basic extra information learned using RF processing, namely, variable importance, partial dependency, and the proximity matrix (Breiman, 2002). Variable importance plots (VIPs) inform about the statistical significance of the effect of each descriptive variable on the resulting prediction (Breiman, 2001). VIPs help users identify which features are important for predictions. Partial dependency plots (PDPs) report the marginal effect of a descriptive variable or multiple variables on the output target (Friedman, 2000). They help users explore how predictions respond to input variables' variations. The VIP and PDP aim to depict the relative influence of each variable (or small group of variables) on the predictions but do not consider existing variable interactions and can be biased if the variables are correlated. The proximity matrix (PM) (Liaw and Wiener, 2007) computes the proportion of times every two observations end up in the same terminal node across all the terminal nodes of the RF model. The idea behind this measure is that similar observations should be found more often in the same terminal nodes than in different nodes. The PM can be plotted using multi-dimensional scaling methods to help users discover data clusters and outliers. However, proximity plots often appear similar or irrespective of the data, raising uncertainty about their usefulness (Hastie et al., 2009).

Post hoc methods generate explanations externally to the RF model. These methods try to mimic the RF predictions to provide an intelligible explanation to users and decision-makers. Post hoc methods are divided into two subcategories: model-agnostic and model-specific. Model-agnostic methods treat the original model as a black box and are designed to be generally applicable (Angelov et al., 2021; Dieber and Kirrane, 2020). In contrast, model-specific methods take advantage of the intrinsic architecture of the original model (Angelov et al., 2021) and can only be applied to a particular type of one or several models. In this survey, we consider an intrinsic model any model that can only be applied to RF or RF and decision trees in general. Only 13% of the surveyed post hoc methods are model-agnostic. Other agnostic methods can probably be applied to RF models. However, the key search used in this survey has limited the results to papers including the term “random forest”. For more examples related to agnostic models, we suggest, for instance, the “Interpretable Machine Learning” book (Molnar, 2022), which describes several agnostic techniques along with application examples, advantages, disadvantages, and implementation references.

#### *4.1.2 Objective*

The objective of interpretability qualifies the extent of the perception provided through explanation. The explanation can enable a global overview, a local explanation, a pattern discovery, or a hybrid objective.

The global overview provides insights into the general behavior of the ML model, such as the most important variables influencing predictions. A global overview that respects human cognitive limitations is the first step to understanding the essential aspects learned by the model. The local explanation focuses on arguing each specific prediction and consists of approximating the original model in a small area around an instance of interest (Angelov et al., 2021). Explaining a specific prediction is challenging, especially if the prediction is subject to moral or legal constraints. The pattern discovery objective aims to unveil patterns and prototypes in the data. For instance, let us consider that we want to predict churn in the telecommunication domain. It would be interesting to discover the prototypes of predicted churners and explain their behaviors to plan focused marketing campaigns. Finally, a hybrid objective combines two or three of the objectives above.

Based on Figure 7, displaying the distribution of the surveyed methods according to the objective south through interpretability, we can see a total overlap of 55%. This result shows that a method can have more than one objective. For instance, rule extraction methods, when model-specific, can enable global overview, local explanation, and pattern discovery. Generating a simple set of rules that approximate the RF model can provide concise information about patterns and prototypes in the data, important variables, and their relationships. For example, in the case of predicting a particular disease (based on information from historical patient data), experts would appreciate discovering a set of rules that explain the different conditions inducing the disease. They would have a global overview of the different prototypes in the data and explain each new prediction depending on the prototype to which it belongs (each rule or reduced set of rules would represent a prototype of patients). They would also be able to understand and analyze each set of conditions and design an adequate treatment accordingly. Based on this analysis, they could continuously improve the predictive model, for example, by adjusting RF hyperparameters or enriching feature inputs.

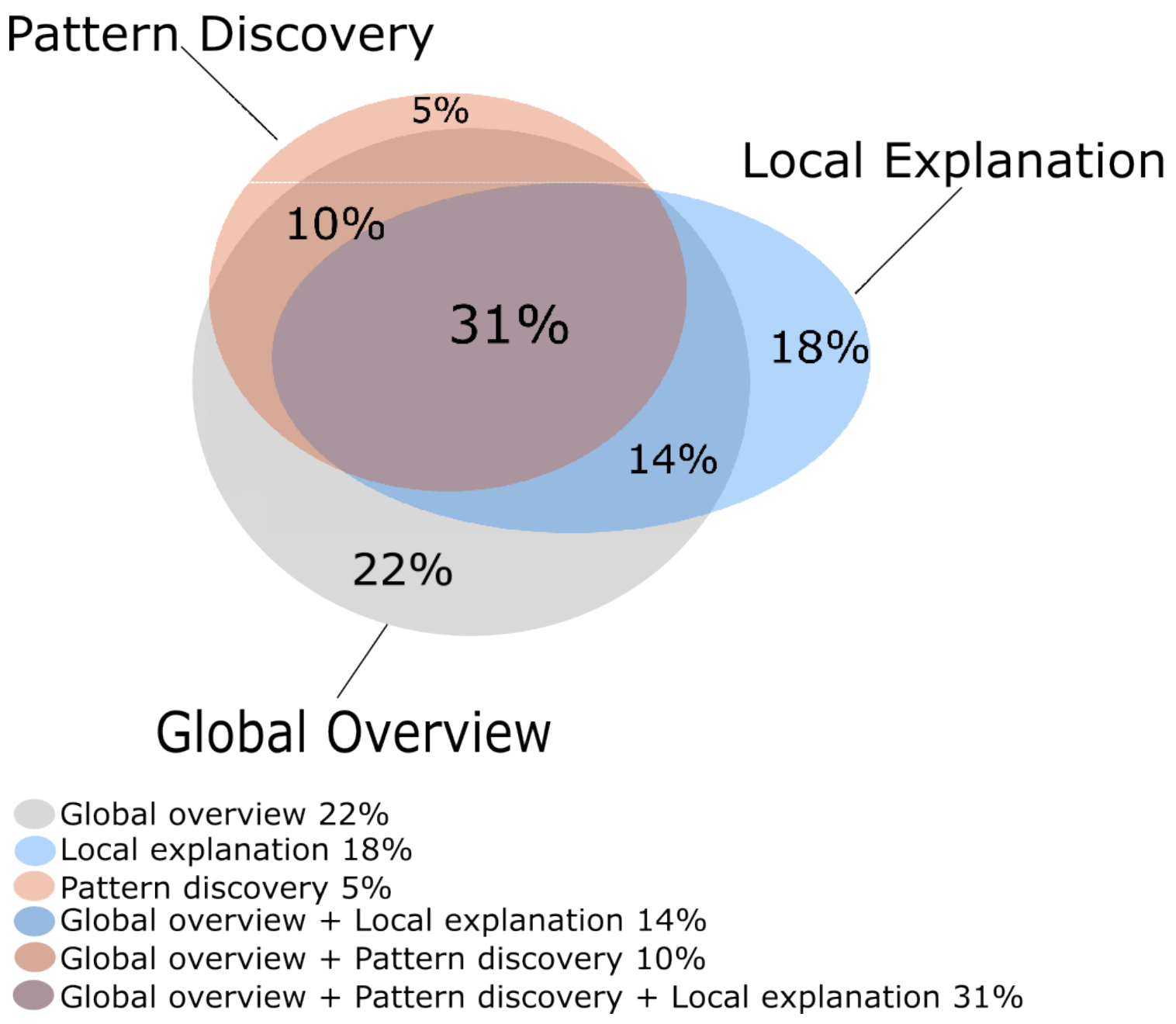

**Figure 7:** Distribution of the surveyed methods according to the objective south through the interpretability of random forest models

*4.1.3 Issue solved and input/output formats*

Based on the 59 selected papers in this review, the issue requiring interpretability is either a regression or a classification problem (binary or multiclass classification). The inputs are continuous, categorical, or mixed data, and the outputs are texts, statistics, or visual graphs. Text outputs concern rule-based explanation, while statistics discuss data distribution, rules metrics, or any numerical information providing insights about the behavior of the RF model. Visual graphs include any plot-based output providing visual explanations, such as scatter plots, box plots, or more sophisticated plots.

Figure 8 illustrates the data flow of the methods interpreting RF, from inputs to outputs passing through the solved issue. The link between two successive nodes is drawn with a width proportional to the flow quantity between the two nodes. The visualization of the different flows does not reveal any evident pattern among inputs, outputs, and solved issues. As shown in this plot, 90% of the surveyed methods process input features that can be either categorical or continuous.

However, several methods use discretization as a preprocessing step to handle continuous variables, such as in (Robnik-Šikonja and Kononenko, 2008), (Ribeiro et al., 2016), (Deng, 2019), and (Debeer and Strobl, 2020). Also, there are more choices for methods interpreting RF classification models than methods interpreting RF regression models. About 88% of the methods can be applied to classification issues, whereas 44% can process regression (by considering hybrid methods). Finally, the outputs' format is distributed in an almost equitable way among the different classes of outputs.

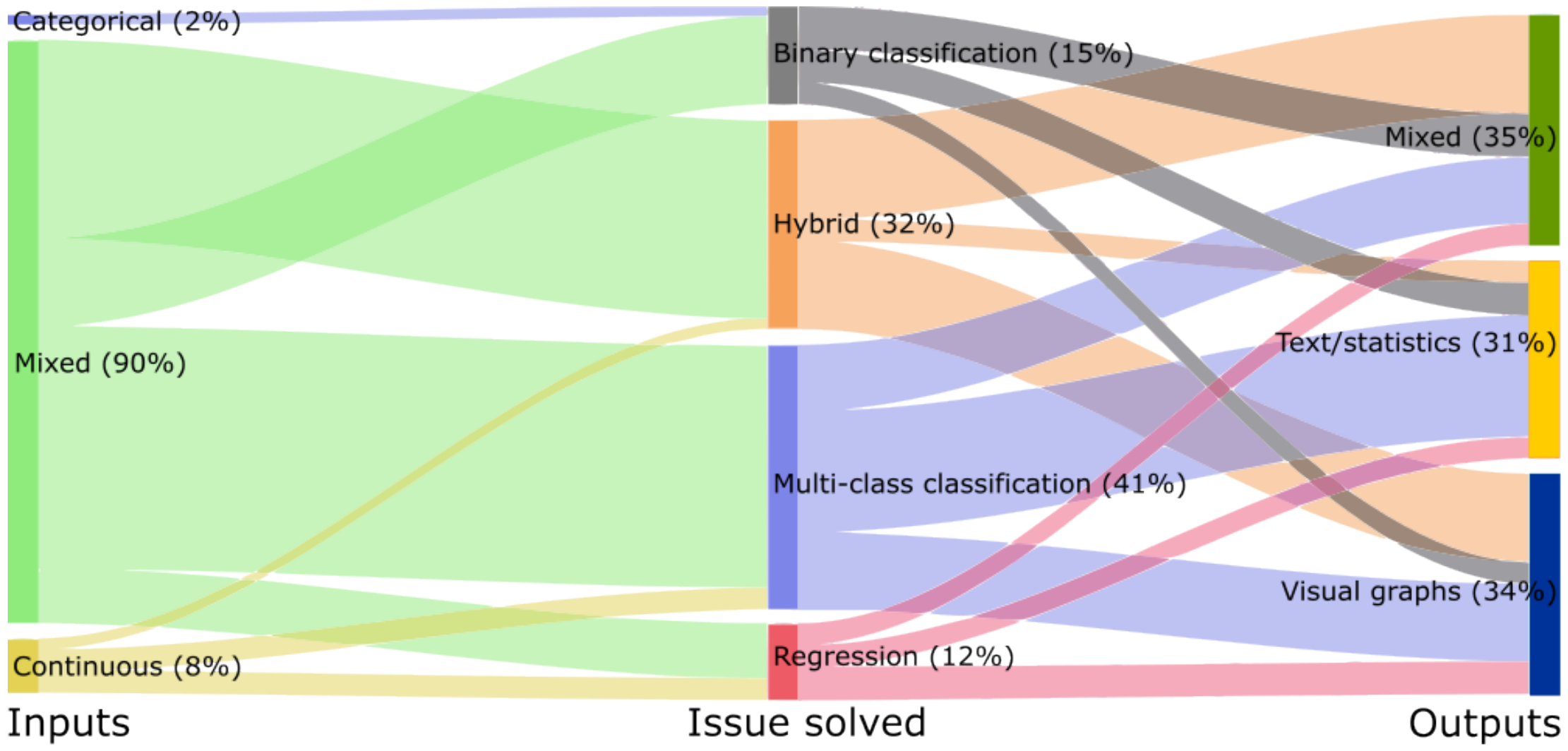

**Figure 8:** Data flow of the surveyed methods interpreting random forest models, from inputs to outputs passing through the solved issue

*4.1.4 Implementation*

The implementation of the proposed methods in the selected papers is mostly done using the R and Python programming languages. Around 48% of the published implementation used Python, 39% used R, and 8% used Java. The remaining 5% used Matlab, the Pyro web application, and the ACE-hipP system. Figure 9 illustrates the evolution of code implementation from 2007 to 2021. It shows that Python was the programming language of choice for developers during the last 5 years.

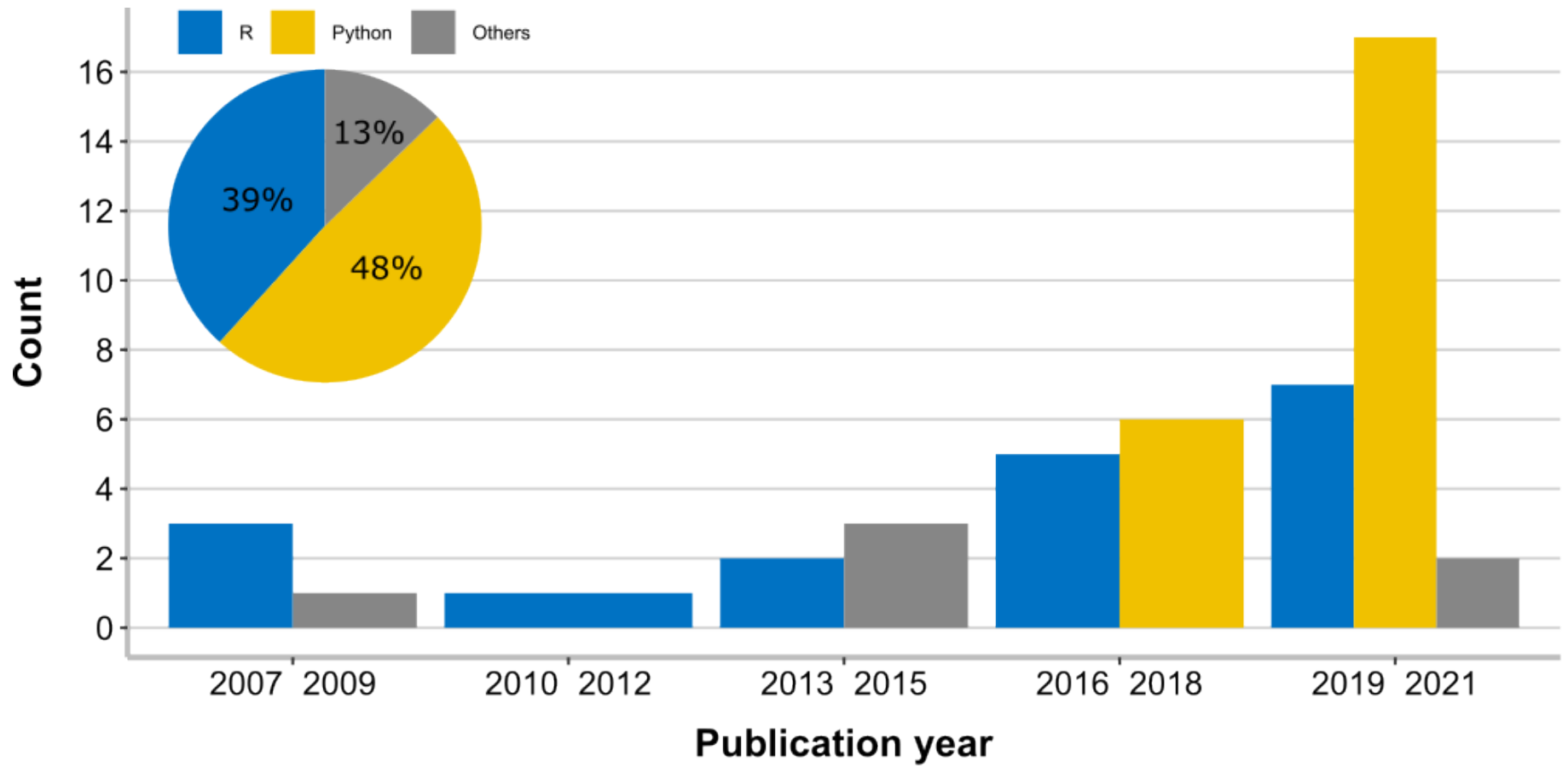

**Figure 9:** Distribution of code implementation of methods interpreting random forest models

*4.1.5 Methodology*

In this section, we classify the selected surveyed approaches based on the methodology used to interpret the RF model. We identified the five following methodologies:

- Size reduction: reduces the number of trees in the forest to produce a simpler model.
- Rule extraction: extracts a representative set of rules for approximating the RF model.
- Features oriented: explains how changing the values of the descriptive features affects the predictions.
- Sample similarity-based: looks at the similarity among instances to explain predictions.
- Template/web interface: framework or web interface developed to facilitate the exploration and explanation of the RF model.
- Hybrid methods: can combine more than one method from the methodologies above.

Figure 10 reports the distribution of the surveyed methods according to the methodology used and the objective of interpretation. This figure shows that almost 40% of the methods use a rule extraction methodology. Around 60% of them allow a global overview of the model, a local explanation of the predictions, and the discovery of patterns in the data. Feature-oriented methods represent about 30% of the studied approaches and are most often used for a global overview. Around 14% of the surveyed methods propose templates or web interfaces that enable a global overview or the discovery of patterns, while 7% of the methods use a sample similarity-based methodology and are generally focused on pattern discovery. The remaining approaches (less than 10%) use a size reduction or hybrid methodology. The following section describes in more details the methodologies introduced in this section.

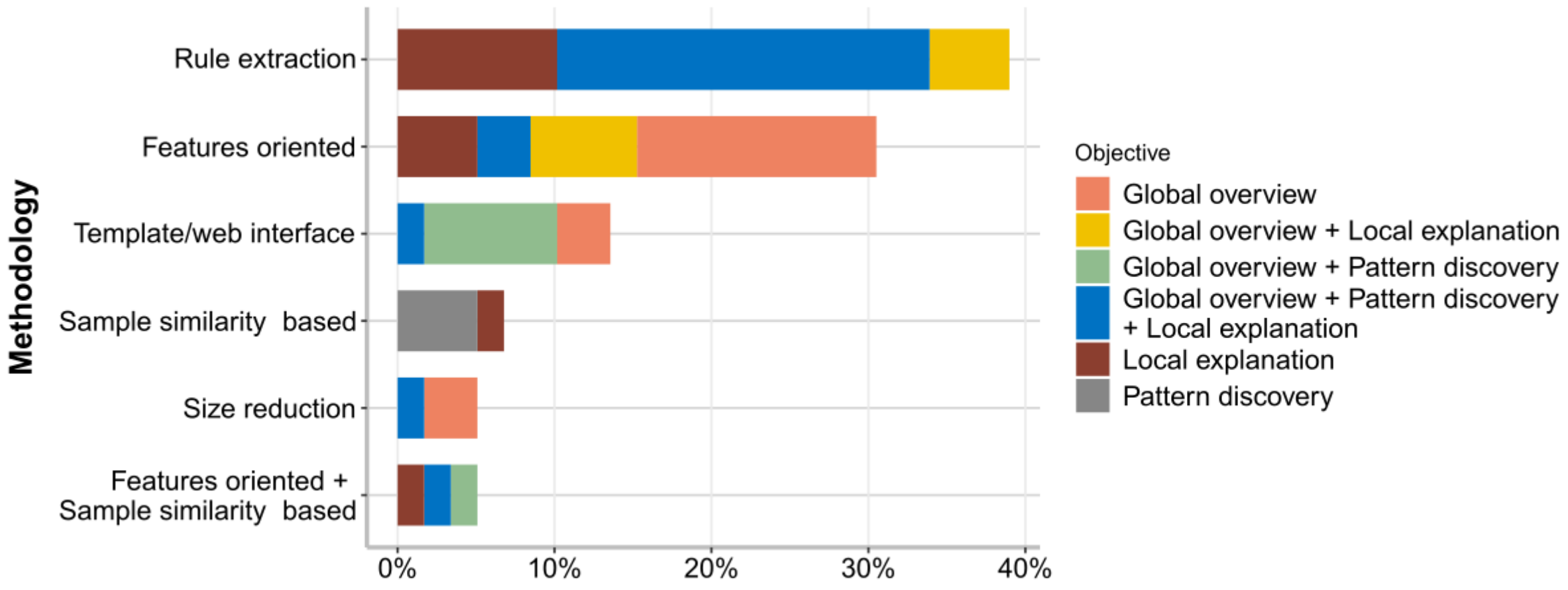

**Figure 10:** Distribution of the surveyed methods according to the methodology used and the objective of interpretation

### 4.2 Review of the surveyed papers according to the interpretative methodologies used

We present in this section a detailed review of the methods interpreting RF according to the methodologies introduced in Section 4.1.5.

#### *4.2.1 Size reduction*

Size reduction methods aim to build a reduced subset of decision trees that is competitive with a large RF model in its predictive performance. Latinne et al. (2001) reduced the number of trees using the McNemar test of significance on the tree predictions. Bernard et al. (2008) used a sub-optimal classifier selection method to obtain a more performant subset of decision trees. Zhang et al. (2021) considered three measures to assess the prediction performance of the trees and to form a sub-forest that achieves better accuracy than the large RF model. Yang et al. (2012) used a margin optimization-based pruning approach to reduce the RF size while improving its performance. Khan et al. (2016) formed an optimal RF sub-ensemble by considering the individual performance of the trees (based on prediction errors) as well as their collective performance (using the Brier score). Finally, Van Assche and Blockeel (2008) used the class distribution estimated from the different decision trees to build a single decision tree that approximates the decisions made by the initial tree ensemble.

The approaches above generally achieve a predictive performance comparable to RF performance. In addition, reducing the size of the forest would make it easier to explore the tree paths. They, however, could remain black boxes depending on the number and the depth of their tree members.

*4.2.2 Rule extraction*

This class of methods builds a rule ensemble to approximate the RF model predictions while considering the complexity of the rule ensemble. This complexity is generally expressed in terms of the number of rules, their lengths, and their coverage. Rule extraction methodology in model-specific methods consists of extracting rules from the RF trees and reducing their number considering the trade-off between their complexity and predictive performance.

In most papers, the authors expressed a rule as the intersection of the conditions defining a path from a root node to a leaf node in a tree. Different methods were used to tackle the trade-off between predictive performance and complexity. Deng (2019) used a complexity-guided condition selection method to select a compact set of rules. Phung et al. (2015) proposed a greedy approach to generate a set of ranked and weighted rules. Mashayekhi and Gras (2017) applied a hill-climbing technique to produce a rule ensemble significantly reduced in size. Hara and Hayashi (2017) used a bayesian model selection algorithm that optimizes the rule ensemble complexity while maintaining the prediction performance. Adnan and Islam (2017) considered different metrics to select a high-quality rule ensemble. Bénard et al. (2020) selected the most relevant rules from an adapted version of the RF model based on their probability of occurrence. Birbil et al. (2020) applied a mathematical programming approach to minimize the number of selected rules and their total impurity. Zhang et al. (2021) based their selection of rules on logical reasoning, sampling, and optimization.

Otherwise, some authors looked at the other possible paths in the trees (not only those from the root to the leaf nodes). Friedman and Popescu (2008), Meinshausen (2010), and Mashayekhi and Gras (2017) selected from the RF nodes the most important rules predicting the outcomes. In the first paper, the authors selected the set of rules based on averaging a weighted rule list. In the second paper, the authors applied a regularized linear regression, and in the third paper, the authors used a heuristic search method and a sparse group lasso method.

In model agnostic methods, the rule extraction methodology consists of constructing a tree or a rule list that mimics the relationship between the inputs and outputs of the RF model, such as in (Johansson et al., 2012) and (Ribeiro et al., 2016).

The if-then semantic of the rules makes the outputs of the rule extraction methods probably the most comprehensible for natural human thinking, provided that the length and the number of rules are acceptable.

*4.2.3 Features oriented*

This class of methods provides explanations based on feature relevance. We can break down this class of methods into 3 sub-categories: VIPs, Dependency plots (DPs), and local decomposition techniques.

The VIP, constructed for RF, shows the importance of the contribution of each descriptive variable on the model prediction performance (Breiman, 2001). Two basic ways are used to measure variables' importance for RF models (Hastie et al., 2009). The first technique measures how much the accuracy decreases (on average) on the out-of-bag samples when each variable is randomly permuted. Debeer and Strobl (2020) and Loecher (2020) developed revisited versions of this technique to increase the interpretability and stability of the outputs and decrease the required execution time. The second approach reports the accumulated improvement in the split criterion (using measures such as the Gini index or entropy) attributed to each splitting variable (Hastie et al., 2009). In recent works, alternative methods were proposed to measure the importance of the variables on the predictive performance. For Hur et al. (2017), Utkin and Konstantinov (2021), and Lundberg et al. (2018), the importance of the features was assessed using the Shapley feature importance measure from game theory, which distributes the model performance among the variables according to their marginal contribution. Vilone and Longo (2020) and Epifanio (2017) proposed a different perspective called the intervention in prediction measure (IPM) to assess variable importance for RFs. IPM computation does not require knowing the prediction performance; it only exploits the structure of the trees forming the RF model. The IPM is a case-wise technique. For each instance, the percentage of times a variable is used in the prediction of the instance is calculated (for each tree). Then, the IPM is averaged for each class in the case of RF classification or globally. Conversely, Ishwaran et al. (2010) and Seifert et al. (2019) proposed another way to assess the importance of the features. The minimal depth (MD) variable importance score of a variable *A* is defined as the average level of the first split based on the variable *A* across all the RF trees with at least one split based on the variable *A*. The surrogate minimal depth (SMD) variable importance is an extension of MD variable importance. It applies the definition of MD not only according to the first appearance in RF splits. It computes surrogate variables for RF and applies the definition of MD to RF primary splits and surrogate splits.

DPs offer another perspective for analyzing the influence of the descriptive variables on the model predictions. PDPs inform users about the global relationship between a feature (or a set of features) and the predicted outcomes. PDPs show the marginal effect of an input variable (or a set of input variables) on the prediction of the fitted model (Friedman, 2000). Parr and Wilson (2021) proposed a revisited approach that computes partial dependencies directly from the training data without relying on the predictions of the fitted model. Their work is motivated by the claim that the same partial dependency algorithm can provide different shapes for different supervised models, which makes it difficult to differentiate between model artifacts and true relationships in the data.

The key disadvantage of PDPs is that they only capture the main effect of the features and ignore possible feature correlations. SHAP DPs (Lundberg et al., 2018) are an alternative solution to the traditional PDPs that better capture interaction effects among variables.

The local decomposition of a prediction by feature relevance is another useful way used to inspect specific predictions. The feature contribution (Palczewska et al., 2014) and forest floor (Welling et al., 2016) methods quantify the contribution of each descriptive variable in each instance prediction, for random forests. Forest floor also provides graphical visuals of the prediction decomposition in a 2D-3D features space, which enables exploring the fitted model structure in a 2D or 3D space.

#### *4.2.4 Sample similarity-based*

Sample similarity-based methods look at the similarity among instances when providing explanations. The proximity matrix inherent to the RF model reports the proportion of time; every two observations end up in the same terminal nodes across all the terminal nodes of the RF trees. The motivation behind this measure is that similar observations should belong to the same terminal nodes more often than dissimilar ones. The proximity matrix can be projected into a 2D space using multi-dimensional scaling (MDS) methods to identify data clusters and outliers (Liaw and Wiener, 2007). Płoński and Zaremba (2014) presented an alternative method that builds supervised self-organizing maps (SOMs) based on the RF proximity matrix (for classification). Their experiments revealed that visualizing the RF proximity matrix with a SOM offers a better understanding of the relationships in the data than MDS visualization. In addition, the SOM learned using the RF proximity matrix achieved better predictive performance than a SOM learned with Euclidean distances. In the same vein, Tan et al. (2020) introduced an adaptive prototype selection method to reveal prototypes according to a distance function derived from the RF proximity. On the other hand, Lundberg et al. (2018) proposed to find clusters in the data based on a supervised clustering of the SHAP feature attributions, which means that the instances are clustered based on explanation similarity.

While the methods above aim for a global overview, other methods consider instance similarities for a local explanation. According to Ribeiro et al. (2016), the local interpretable model-agnostic explanations (LIME) model aims to locally approximate a black box-model by an interpretable model (linear regression as an example). First, LIME creates synthetic samples in the neighborhood of the instance to explain. Next, it computes a similarity index between this instance and the sampled instances. It then fits the original and interpretable models to these samples. Lastly, LIME minimizes a loss function (which measures the loss in prediction between the original model and an interpretable model) weighted by the proximity to the instance being explained, plus a complexity term (which measures the complexity of the interpretable model). The best interpretable model is the solution to this optimization problem.

On the other hand, Iforest (Zhao et al., 2019), a model-specific method, explores the similarities among the decision paths that generate the same predictions in the RF trees. It uses the t-distributed stochastic neighbor embedding (t-SNE) technique to project all the decision paths onto a 2D space so that users can easily visualize and examine the properties of path clusters that induced the decision of a specific input instance. According to Brophy and Lowd (2020), the tree-ensemble representer-point explanations (TREX) method uses the structure of the trees to build a tree ensemble kernel, which acts as a similarity measure among instances. By using this kernel in a surrogate model that approximates the original tree ensemble, TREX computes the global or local importance of each training example in the prediction of each specific instance, resulting in an instance-attribution explanation of positive and negative training points. Finally, Blanchart (2021) proposed to characterize the model into pure decision regions (regions over which the model makes a constant prediction over all classes) under the form of a collection of multi-dimensional intervals. Using counterfactual (CF) reasoning, they determine the closest CF example associated with a specified instance and the geometrical characterization of its decision region. The CF explanation describes the slightest change to the feature values that changes the prediction value to another predefined value.

#### 4.2.5 Template/web interface

This class of approaches covers methods that provide web interfaces or templates ready to apply to interpret RF predictions based on existing published approaches. Such tools should help researchers easily apply/monitor the implemented methodologies to their research problems. Many researchers provided interactive visualization tools based on the existing methodologies, allowing the users to manipulate the visible results via input devices such as a keyboard or mouse.

The randomForestExplainer package (Paluszy, 2017) computes and plots various statistics measuring the importance of variables and their interactions in RF models. Although the package uses existing methodologies (except for some simple extensions), it provides various graphical capabilities that facilitate visualizing and interpreting the results. The ggRandomForests (Ehrlinger, 2015) package is also devoted to visually understanding RF models. It allows for examining variable importance and minimal depth measures. It also enables investigating variable associations with the variable dependencies plots or using minimal depth interaction plots. Along the way, the package offers commands for modifying and customizing results. Similarly, the edarf (Exploratory Data Analysis using Random Forests) (Jones and Linder, 2016) package contains useful functions for computing and visualizing the partial dependencies of features, the permutation importance of covariates, and the proximity between data points according to the fitted model. In the same vein, Smith and Alvarez (2021) published a template for computing and visualizing Shapely values for binary classification problems.

Alternatively, Quach (2012) developed the irfplot package (interactive Random Forest Plot) and combined it with the iPlots eXtreme tool (Urbanek, 2011) to offer an interactive visualization of parallel variable coordinates, variable importance, and proximities plots. Rfviz is another sophisticated interactive visualization tool designed for interpreting RF predictions in a user-friendly way. The Rfviz tool allows visualizing and interacting with the parallel coordinate plot of the descriptive variables, the parallel coordinate plot of the variable importance scores, and the rotational scatterplot of the proximities.

Some tools are devoted to specific domains of applications. Intended for processing electronic health records (EHRs), the VERONICA (Rostamzadeh et al., 2021) system incorporates a visual analytic module that utilizes the natural classification of features in EHRs to identify the group of features with the strongest predictive power, integrating different sampling strategies, supervised ML techniques (including RF), analytics algorithms, visualization techniques, and human-data interaction.

While the tools above generally employ already published methodologies, their main advantage is that they help researchers and users easily and effectively apply/monitor the implemented methodologies to their research problems.

Figure 11 classifies the different methods listed in Table 1 according to the objective of interpretability, the methodology used, and the issue solved. The detailed taxonomy of the 59 studied methods can be found on GitHub (see Section Declarations). This figure offers to users a visual tool for the identification/selection of the appropriate interpretative methods given the objective of interpretability, the methodology used, and the nature of the issue under study.

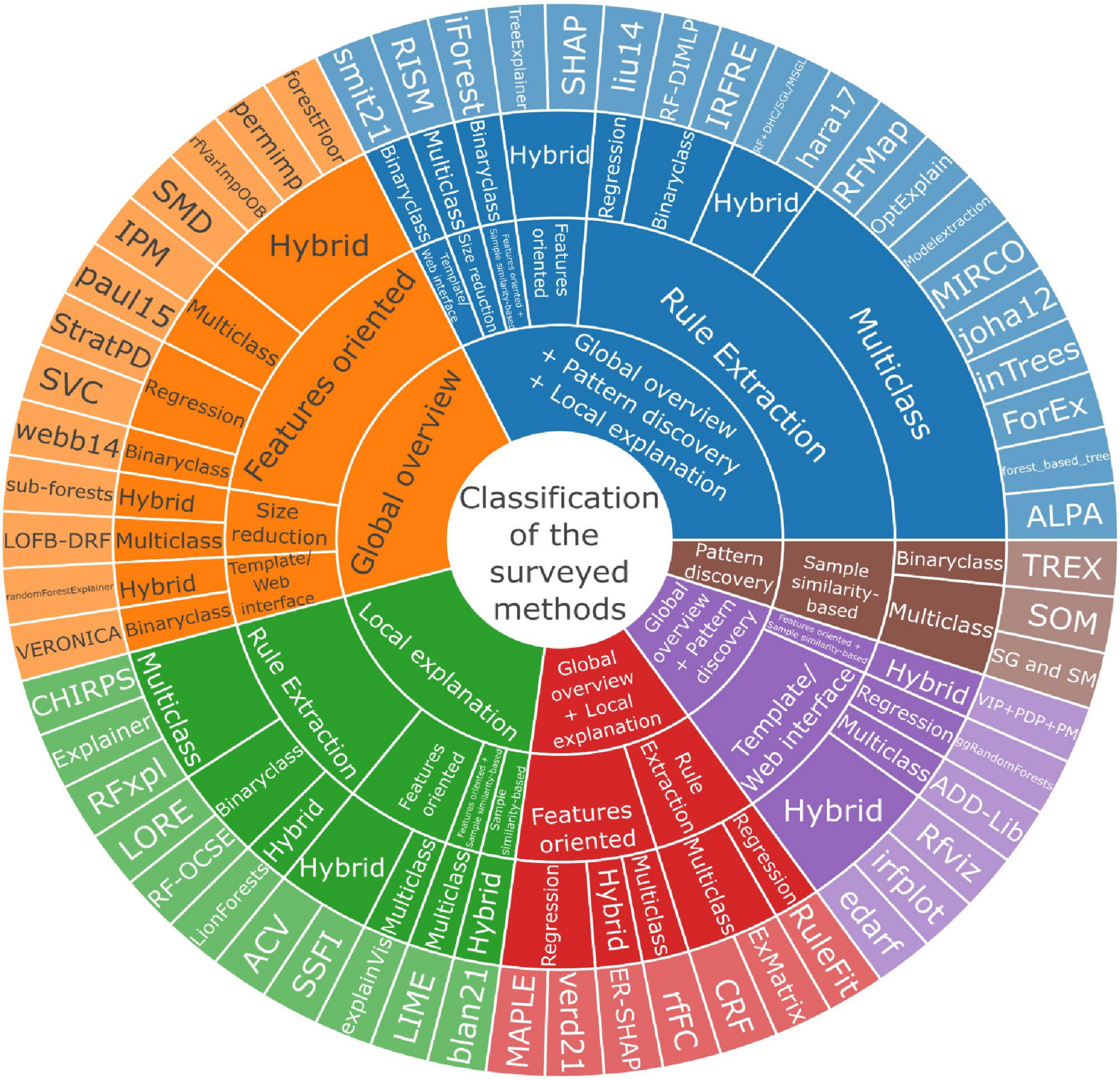

**Figure 11:** Classification of the surveyed methods according to the objective of interpretability, the methodology used, and the issue solved

In practice, it is recommended to experiment with various interpretative methods, which should ideally lead to similar or complementary conclusions. Moreover, the provided explanations should be adapted to users' cognitive capacities and preferences (Haddouchi and Berrado, 2018; Kovalerchuk et al., 2020; Lahav et al., 2019). In an ML project applied to a specialized domain, explanations may differ depending on the user trying to understand the predictive model (developer, data analyst, technical user, domain specialist, lay users, etc.).

The usability of the provided explanations should be validated before deploying an RF model. Usability experiments should verify that the presented RF model is accepted for deployment based on the diverse explanations examined by the different users (including the team project and the end-users). Alternatively, it may occur that a user rejects the model due to misleading explanations (Lipton, 2018). Such deficient explanations, also called quasi-explanations (Kovalerchuk et al., 2020), are generally a result of using concepts or terms foreign to the user domain or scientific background.

We investigated the inclusion of usability studies in the surveyed papers. We provide in the next section the obtained results. We also present the expected advantages and disadvantages of the different methodologies used to interpret RF based on the usability studies reported in the surveyed papers and several other papers tackling the usability issue.

## 5. USABILITY STUDIES

### 5.1 Importance of usability studies

It is important to remember that ML interpretation is not a machine activity; it fundamentally involves human perception and interaction (Kovalerchuk et al., 2020). As related by Lahav et al. (2019) and Poursabzi-Sangdeh et al. (2021), many ML interpretability studies focus on developing methods producing explanations that are assumed to be easily understandable to humans. Through experimental studies, these works revealed that there is a divergence between ML interpretability designers' expectations and the practical results in real-world applications.

In the same vein, Kovalerchuk et al. (2020) provided many examples of explanations that can be easily rejected by users without a background in ML. Explanations based on weights, distances, layers, or deep trees could be meaningless in real-world applications. For instance, a weighted summation of the blood pressure and temperature is foreign to the medical language. Also, a deep tree or a large ordered rule list could be difficult to trace and summarize by a clinician, especially if they use non-conventional terms or split values (Huysmans et al., 2011).

Moreover, some visualizations that are assumed to facilitate understandability and communication can instead decrease the usability of the explanations to end-users. Experiments with users not related to the field of data science showed that there is no evidence that visualizations increase the level of perceived ML model usability compared to the absence of visualizations (Haas, 2021).

Thus, designing meaningful explanations for users is challenging. It should consider users' backgrounds, knowledge, and preferences, which can vary considerably (Nourani et al., 2018; Ali et al., 2023; Molnar, 2022).

It is, therefore, necessary to validate the designer's assumptions about the usability of an interpretative model based on interactive experiments with users. The designer should verify that interpretations are acceptable by the different users, leading to the deployment of the model in practice (Lahav et al., 2019).

It also may be interesting to consider dynamic formats of explanations based on users' experimental feedback (Nourani et al., 2018) because although the main objective of users' experiments is to produce meaningful explanations, different sub-objectives should be defined in consideration of the different classes of users and applications (Ali et al., 2023).

Limited work on RF interpretative methods has reported usability studies. Out of the 59 surveyed works, only eight papers report usability experiments. Their authors proposed different testing protocols based on subjective and objective measures for quantifying the understandability and utility of the provided explanations. The experiment participants were either human subjects on Amazon Mechanical Turk without a background in ML or users familiar with the ML field.

### 5.2 Usability studies related to RF interpretative methods

Zhao et al. (2019), Neto and Paulovich (2021), Mazumdar et al. (2021), and Bastani et al. (2017) performed user studies over respectively 10, 13, 15, and 46 participants with a background in ML. Their objective was to verify that their model explanations can help technical users, such as data scientists, understand and validate the predictions of an RF model.

Alternatively, Lundberg et al. (2019), Lundberg et al. (2018), and Tan et al. (2020) aimed to validate

through user studies, including respectively, 33, 34, and 42 Amazon Mechanical Turk without a background in ML, that their methods allow the best agreement with human intuition in predicting outcomes compared to other existing methods.

Finally, Ribeiro et al. (2016) evaluated their agnostic method with both technical and non-technical human subjects, by setting three main objectives. The authors recruited a hundred of Amazon Mechanical Turk without a background in ML to check the two first objectives. They first verified that the provided explanations can help users decide from a set of classifiers, which classifier generalizes better. They also verified that their explanations can help non-expert users improve the predictive model by modifying the set of attributes used in the model. The third objective required users familiar with the ML domain (27 graduate students with knowledge in ML). The authors aimed to demonstrate that explaining individual predictions helps users know when and why the model will likely fail in predicting outputs. It should be noted that the experiments with real users in (Ribeiro et al., 2016), were conducted using a set of classifiers that, unfortunately did not include the RF model. Since the experimented model is an agnostic model (that could be applied to any classifier), the conclusions are assumed to be generalizable to other classifiers. However, The findings should be verified for other classifiers and real-world applications.

Based on the different usability works reported in this paper, it is clear that user studies depend on their designers' choices. The designers' choices regarding the targeted audience, number of users, study objectives, predictive issue nature, data complexity, and questionnaires' content and score assignment vary greatly over the different studies. Although the literature proposes various user experiment designs, there is still no consensus about how the usability of the explanations can be assessed effectively (Ali et al., 2023; Saeed and Omlin, 2023), especially for end-users without a background in ML.

Liao et al. (2020) investigated the explanation needs of end-users. They proposed a Question Bank (QB), including a set of prototypical questions that should help designers identify and address explanation needs. They also mapped the proposed questions to explanation methods available in the literature. Sipos et al. (2023) studied the applicability of this QB in a specific usage context. They proposed an extension of this QB with other questions. They also enriched the existing questions, aiming that their work will provide a basis for future studies on the identification of explanation needs in different contexts. In real-world ML projects, the authors suggested using the QB as a starting point and customizing it based on team discussions.

In the same vein, Jin et al. (2022) suggested considering users' feedback while designing post-hoc explanations. The authors conceived the EUCA framework, which provides a prototyping protocol (including tools and methodologies) that can guide the creation of solutions that consider the context-specific explainability needs. They listed several end-user-friendly explanatory forms that can be used as building blocks in a participative design, including technical and non-technical users. They also detailed these explanatory forms based on several criteria (such as their applicable explanation goals, visual representations, advantages and disadvantages, and examples of algorithmic implementations), and provided supporting material for their practical use (including prototyping cards, templates, and examples of application). While the EUCA protocol includes a limited number of explanatory forms and lacks the quantitative evaluation of usability, it provides a practical protocol example for designing feasible collaborative solutions seeking explainability.

User studies in the literature employ diverse metrics to quantify usability. We refer to (Rong et al., 2022) work, which presents a survey of metrics used in recent works, hence providing ideas on how to collect and evaluate human feedback on explanations. The authors in this work deeply analyzed several user studies' experimental designs, participants, and measures. They also proposed a general guideline for human-centered user studies.

Works such as (Liao et al., 2020), (Sipos et al., 2023), (Jin et al., 2022), (Rong et al., 2022), and (Molnar, 2022) can be used as starting points for the design of practical solutions for interpreting RF models, including diverse explanatory forms. Our survey of methods interpreting the RF model is not exhaustive but provides a good basis for the diverse exploratory forms available for the RF model. We summarize in Table 2 the expected advantages and pitfalls of the methodologies presented in Section 4.2, based on the different works addressing the usability topic reported in this paper. This table along with the taxonomy presented in this paper and the detailed classification provided on GitHub (see Section Declarations), could be used as supporting material for designing a user-friendly solution that considers the different roles, needs, and preferences of the users requiring explanations.

**Table 2:** Advantages and disadvantages of the RF interpretative methodologies

| Methodologies | Description | Advantages | Disadvantages |
|---|---|---|---|
| Size reduction methods | Build a reduced subset of decision trees that is competitive with a large RF model in its predictive performance. | Reducing the size of the forest would make it easier to explore the tree paths for a local explanation. | The subset of decision trees generally remains a black box depending on the number and the depth of its tree members. |
| Rule extraction methods | Build a rule ensemble that approximates the RF model predictions while considering the complexity of the rule ensemble. This complexity is generally expressed in terms of the number of rules, lengths, and coverage. | The if-then semantic of the rules is intuitive for human thinking (if rules are built with intelligible features)<br>Rules facilitate communication about the model predictions.<br>Rules can allow for local explanation or data clustering, where each cluster is explained by one or a few rules.<br>Rules inform about interactions between features. | Determining the best tradeoff between the complexity of the rule ensemble and the predictive performance is challenging (tackled differently among papers) and generally time-consuming. It is sometimes not aligned with user's needs.<br>The process of building the rule ensemble does not consider the meaningfulness of the features and splits employed. In addition, by choosing the least complex rules, this process generates sometimes counter-intuitive rules for domain experts.<br>Rule ensemble building process often gives more importance to the improvement of the predictive performance compared to the interpretability coverage (many instances remain unexplained).<br>The rule ensemble is sometimes made up of multiple overlapping rules, which can blur the overall overview of the final model. |
| Feature-oriented methods | Provide explanations based on feature relevance. Can be divided into three sub-categories: VIPs, DPs, and Local decomposition. | VIPs help users identify which features are globally important for predictions.<br>DPs help users explore how predictions respond to input variable variations.<br>Local decomposition based on feature relevance helps users to understand and compare individual predictions. | PDPs and VIPs do not consider variable interactions and can be biased if the variables are correlated.<br>SHAP DPs better capture interaction effects among variables but are generally more computationally consuming.<br>The realistic number of features to represent in PDPs is limited. It is hard for Humans to imagine more than three dimensions.<br>Local decomposition (such as the summation of weighed heterogeneous features) could be not meaningful and foreign to the application domain language. |
| Sample similarity-based *methods* | Look at the similarity among instances when providing explanations. | Visualizations based on the RF proximity matrix or SHAP explanation similarity allow the identification of data clusters and outliers.<br>Local explanations based on similarity (similar or counterfactual examples) help verify and compare decisions. | Visualizations based on RF proximity often appear similar or irrespective of the data, raising uncertainty about their usefulness.<br>Similar or contrafactual examples could cause confusion if they are counter-intuitive or poorly described (via non-meaningful features). |
| Template/web interfaces | Cover methods that provide web interfaces or templates ready to apply to interpret RF predictions based on existing approaches. | Help researchers and users easily apply/monitor existing methodologies to their research problems.<br>Allow users to easily manipulate the visible results via input devices such as a keyboard or mouse. | Interactive and sophisticated visualizations generally require supplementary text explanations and tutorials.<br>Non-conventional visualization forms can be misleading and often require extra effort for users compared to text or tables. |

## 6. SUMMARY AND CONCLUSIONS

As reported in this paper, RF has the advantage of being an intelligible algorithm regarding its building process. However, gaining visibility over the entire process that induces the final decisions by exploring each RF decision tree is complicated, if not impossible. This complexity limits the acceptance and deployment of RF models in several fields, such as health care, biology, and security. The interpretability of RF models within these fields is required because the more transparency the model gains, the more users will trust it and take action based on its results. Otherwise, if the decision-making process is not clearly explained and understood, errors may occur and go unnoticed until they cause technical, financial, legal, moral, or ethical damage that may be difficult or impossible to repair. Thus, more and more approaches are proposed in the literature to interpret RF models, especially in the last 3 years.

This paper provides a literature review of methods interpreting RF models, following a systematic methodology. This review presented at first a bibliographic analysis of the surveyed papers through the analysis of the number of publications, the author's literature map, and the terms' co-occurrence map. It then proposed a taxonomy and a distribution analysis of the surveyed methods based on the main properties that could help users choose the most suitable methods for the issue under study. The taxonomy criteria include the stage of explanation, the objective of interpretability, the type of the problem solved, its input and output formats, the methodology used for providing explanations, and the programming language used in the implementation of the method.

The analysis of the surveyed methods showed that around 97% of the papers propose post hoc methods (3% ante hoc). About 87% of them are model-specific methods, thus taking advantage of the intrinsic architecture of RFs or decision trees in general. The remaining 13% are model-agnostic. This tendency suggests that the RF predictive structure enables the discovery of diverse kinds of explanations that are worthwhile to investigate.

The analysis also revealed that about 90% of the surveyed methods can process mixed data (categorical and continuous inputs). It showed, in addition, that the output format of explanations varies equitably among texts, statistics, or visual graphs format.

The distribution of the surveyed methods depending on the nature of the issue solved showed that there is more choice for methods that can be applied to classification issues (around 88%) than for methods that can be used for regression (around 44%). In addition, it has been noticed that Python is the programming language of choice used to develop the surveyed methods during the last five years. The comparison of the surveyed methods based on the objective of interpretability (global overview, local explanation, pattern discovery, or a hybrid objective) exhibited about 55% of overlap, meaning that most methods enable more than one objective.

We have analyzed the methodologies used to interpret the RF model and structured them into five classes. Size reduction methods reduce the number of RF trees to produce a simpler model. Rule extraction methods extract a representative set of rules for approximating the RF model. Features-oriented approaches explain how changes in the values of the descriptive features affect the predictions. Sample similarity-based methods investigate the similarity among instances to explain predictions. Finally, template/web interface methods offer frameworks or web interfaces to facilitate the exploration and explanation of the RF model using already published methodologies. Hybrid methods can combine more than one method from the methodologies above.

We have shown that almost 40% of the methods use a rule extraction methodology, and that around 60% of them allow a global overview of the model, a local explanation of the predictions, and the discovery of patterns in the data. Feature-oriented methods represent about 30% of the studied approaches and are most often used for a global overview. Around 14% of the surveyed methods propose templates or web interfaces that enable a global overview or the discovery of patterns, while 7% of the methods use a sample similarity-based methodology and are generally focused on pattern discovery. The remaining approaches (less than 10%) use a size reduction or hybrid methodology.

We have provided a visual tool for the identification/selection of appropriate interpretive methods given the interpretability objective, the methodology used, and the nature of the problem under consideration.

We recommend, in practice, comparing and combining different methods of interpretation to extract various explanations, which should ideally lead to similar or complementary conclusions. These conclusions should be validated based on rigorous usability studies with technical and non-technical users. We reported in this work that only around 13% of the surveyed methods presented usability studies, suggesting that the field of RF interpretability is still far from the level of the required scientific rigor. We thus reported several works tackling the usability issue, aiming that they could help design practical end-user solutions that gather diverse RF explanatory forms.

In future work, we plan to deeply analyze the literature related to usability studies in ML. We are interested in designing a practical protocol and toolbox that could be used in usability experiments. These tools are aimed at guiding the choice of the most effective solutions for interpreting RF models depending on the problem under study, the concerned users, and the objectives of interpretability. We also plan to focus on interpreting the RF model via the rule extraction methodology. We consider this class of methods a key to efficient and natural understanding and communication about the model learned by RFs. As reported in this survey, rule extraction methods can allow for a global overview of the model, a local explanation of the predictions, and the discovery of patterns. The main challenge for this kind of method is to find the best trade-off between the complexity and the predictive performance of the final rule ensemble. We intend in future work to analyze the existing rule extraction methods to figure out the main characteristics that come into play when addressing this trade-off. We aim to determine the existing shortcomings and propose possible solutions to resolve them.

## DECLARATIONS

### Funding Information

The authors did not receive support from any organization for the submitted work.

### Conflict of interest/Competing interests

On behalf of all authors, the corresponding author states that there is no conflict of interest.

### Code availability

The implementation and the computational work are done using the R language and environment for statistical computing. The code and data files used are available via https://github.com/HMAISSAE/RF_SLR.

### Availability of data and materials

All data used to perform this review is available via https://github.com/HMAISSAE/RF_SLR.